\documentclass[preprint,12pt]{elsarticle}

\usepackage{graphicx}
\usepackage{amssymb}
\usepackage{amsmath}
\usepackage{lineno}
\usepackage{amsfonts}
\usepackage{amsmath}
\usepackage{algorithm}
\usepackage{algorithmic}
\usepackage{float}
\usepackage{subfigure}
\usepackage{subcaption}
\usepackage{booktabs}
\usepackage{tabularx}
\usepackage{caption}
\usepackage{multirow}
\usepackage{makecell}
\usepackage{microtype}
\usepackage{array}

\renewcommand\tabularxcolumn[1]{m{#1}}

\newcolumntype{C}[1]{>{\centering\arraybackslash}m{#1}}
\newcolumntype{Y}{>{\centering\arraybackslash}X}

\journal{Neurocomputing}

\begin{document}
\begin{frontmatter}

\title{IG-GAN: A Generative Adversarial Network for Aerodynamic Data Generation Based on Intrinsic Geometry} 

\author[inst1]{Ying Yan}
\ead{yingyan@stu.hebust.edu.cn}

\author[inst1]{Liwei Hu\corref{cor1}}
\ead{liweihu@hebust.edu.cn}

\author[inst1]{Xiaoming Zhang\corref{cor1}}
\ead{zhangxiaom@hebust.edu.cn}

\cortext[cor1]{Corresponding authors.}

\affiliation[inst1]{
    organization={School of Information Science and Engineering, Hebei University of Science and Technology},
    city={Shijiazhuang},
    postcode={050091},
    country={China}
}

\tnotetext[t1]{This work is Funded by Science Research Project of Hebei Education Department. (No. QN2026857)}

\begin{abstract}
Existing generative models learn data distributions in flat Euclidean space. 
However, most data in our real world are manifolds embedded in high dimensional Euclidean space. 
Therefore, we propose an intrinsic-geometry-based generative adversarial network (IG-GAN) for data generation in the field of aerodynamics. 
The generator of the IG-GAN represents aerodynamic data as a piecewise smooth manifold constructed by Bézier surfaces, and the generator tries to learn the coefficients of each Bézier surface to further combine multiple Bézier surfaces into a smooth manifold automatically. 
The discriminator in the IG-GAN is a radial-basis-function based discriminator (RBF-D). 
Experimental results show that IG-GAN achieves lower predicted Mean Squared Errors (MSEs) than those of three baselines. 
Specifically, on the Burgers' equation dataset, IG-GAN reduces the predicted MSE of velocity $u$ by $97.41\%$ compared with state of the art SSL-Transformer. 
Additionally, on the ONERA M6 aircraft dataset, IG-GAN reduces the overall MSE of nine aerodynamic coefficients by $82.95\%$ compared with SSL-Transformer.
\end{abstract}


\end{frontmatter}


\section{INTRODUCTION}
\label{section_introduction}
Existing generative methods, e.g. generative adversarial networks (GANs) \cite{hu2022flow}, diffusion models \cite{croitoru2023diffusion}, and transformers \cite{amatriain2023transformer}, perform data generation under flat Euclidean spaces \cite{li2023euclidean, kim2025novel}. 
Using these methods to generate data is essentially a way to learn the geometric structure of the data (or data distribution). 
Data generation under flat Euclidean space contains a hybrid representation of data geometry, i.e. these methods learn not only the intrinsic geometry (e.g., curvature and torsion \cite{hu2025learning})  of the data itself, but also the extrinsic geometry (e.g. the specific coordinate values \cite{fang2024intrinsic}, normal vector \cite{koestler2022intrinsic}, etc. ) determined by the external Cartesian coordinate system. 
However, most real-world data are low-dimensional embeddings with complex and curved geometries in high-dimensional space, which implies that those geometries of these data are governed by their intrinsic geometric properties, independent of their position in the external coordinate system \cite{meila2024manifold}. 

Recent studies that use intrinsic geometry-based methods to represent objects have demonstrated advantages in capturing the inherent structural properties of objects in high-dimensional space \cite{zhang2023Data-Informed}. 
For example, a hyperbolic manifold can be used to represent data with tree-like geometry \cite{mettes2024hyperbolic}, and a sphere manifold can be used to approximate the hyper surface induced by deep learning loss functions \cite{hu2026geodesic}. 
In aerodynamics, there is a growing trend in intrinsic geometry representations for aircraft. 
Xiang et al. \cite{xiang2023manifold} used a set of Bézier curves to represent the geometry of UIUC airfoils,  and further calculated the intrinsic geometric features to reduce the coefficient of drag ($C_D$) prediction errors. 
Based on the study of Xiang, Hu et al. used a set of Bézier surfaces to represent the intrinsic geometry of DLR-F11 wing in 3-D space, and using connection and curvature to reduce the coefficient of pressure ($C_P$) prediction error on the wing surface \cite{hu2025learning}. 
The above studies demonstrated that the intrinsic geometry methods can be used to accurately represent the aircraft geometry and further to reduce the prediction errors of coefficients. 
However, accurately approximating arbitrary given data geometries still requires a large number of Bézier curves or surfaces. 
As calculating the coefficients for each curve or surface is computationally expensive, automatically learning these coefficients via neural networks represents a promising solution.

Motivated by intrinsic geometry methods, we proposed an intrinsic geometry-based GAN (IG-GAN) for aerodynamic data generation. 
The generator of the IG-GAN is used to simulate the process of approximating the data geometries using Bézier curves/surfaces, and the coefficient matrices of each Bézier curves/surfaces are learned by the generator. 
As the data generation ability of the generator improves, the data distinguish ability of the discriminator needs to be improved correspondingly. 
Therefore, we use the RBF-D \cite{hu2022flow} as the discriminator. 
Together, the Bézier-based generator and RBF discriminator enable IG-GAN to learn aerodynamic data distributions from an intrinsic-geometric perspective.

We highlight the contributions of this work as follows.

1) We propose IG-GAN for nonlinear aerodynamic data generation, where the generator learns the coefficient matrix of a Bézier-inspired intrinsic geometric representation.

2) We extend the intrinsic-geometry formulation from a two-dimensional parametric construction to high-dimensional aerodynamic data by using learnable coefficient matrices for manifold reconstruction.

3) We validate IG-GAN on the Burgers' and ONERA M6 datasets. 
Compared with Self-supervised learning based Transformer(SSL-Transformer)\cite{xu2024self}, IG-GAN reduces the test Mean Squared Error (MSE) by approximately $97.41\%$ and $82.95\%$ on the two experiments, respectively.

The rest of this article is organized as follows. 
Section~\ref{section_related_works} reviews related work on intrinsic-geometry-based aerodynamic modeling and generative models in fluid mechanics. 
Section~\ref{section_method} details the mathematical formulation of IG-GAN, including the Bézier-based generator. 
Section~\ref{section_experiments_and_results_analysis} presents two experimental studies. 
Finally, Section~\ref{section_conclusion} concludes and discusses future work.

\section{RELATED WORK}
\label{section_related_works}

\subsection{Intrinsic Geometry-Based Aerodynamic Data Modeling}

In the field of intrinsic geometry, the curved geometry of aerodynamic data (e.g. 3-D aircraft, flow field data, and aerodynamic coefficients) is typically mapped into a low-dimensional space by a homeomorphic function \cite{chen2022learning, van2022effect}, and the aircraft geometry can be expressed by the low-dimensional parameter space and the homeomorphic function.

Wang et al. adopted cubic splines to transform UIUC airfoil geometries from 2-D Cartesian space into a 1-D curvilinear coordinate, and then a variational autoencoder generative adversarial network (VAEGAN) was utilized to generate various airfoils as well as the corresponding aerodynamic coefficients based on the 1-D curvilinear coordinates \cite{wang2021airfoil}. 
Deng et al. utilized a univalent transformation to convert the geometrical information of airfoil mesh from 2-D Cartesian coordinates to curvilinear coordinates, and then the curvilinear coordinates of airfoils are input to a vision transformer to predict transonic flow over the airfoil \cite{deng2023prediction}. 
Although, the curvilinear coordinates are built based on manifold theory, there is no fundamental difference between using curvilinear coordinates and using Cartesian coordinates to represent geometries, because there is no intrinsic geometry features calculated by the homeomorphic function. 

Based on the above studies, Xiang et al. constructed a 1-D manifold for a given 2-D UIUC airfoil by a set of Bézier curves, and further calculated the Riemannian metric on different points of airfoils through differential operations on Bézier curves. The Riemannian geometry, as an intrinsic geometric feature, results in a significant reduction of $C_D$ \cite{xiang2023manifold}. 
Hu et al. constructed 2-D embeddings for DLR-F11 aircraft in 3-D space by a set of Bézier surfaces, and calculated Riemannian connections and curvatures based on the differential of Bézier surfaces, which reduce the predicted errors of $C_P$ distributions on the wing surface \cite{hu2025learning}.

The above studies demonstrated that the intrinsic geometry method has advantages in representing aircraft geometries and in reducing the prediction errors of aerodynamic coefficients. 
However, it is complex to calculate the coefficient of each Bézier surface to further approximate the aircraft geometry, especially at components where the geometry is significantly unflat such as the junction of the fuselage and the wing.

\subsection{Generative Models-Based Nonlinear Aerodynamic Data Modeling}

Generative models provide an alternative paradigm for modeling complex nonlinear aerodynamic data by learning implicit mappings among flight conditions, aircraft geometries, and flow variations. 

Among various generative frameworks, generative adversarial networks (GANs) \cite{goodfellow2014gan} have been widely adopted in aerodynamic modeling due to their capability to capture the distribution of input data. 
Hybrid aerodynamic generative models have incorporated radial basis functions into the GAN framework to improve local approximation capability and training stability, thereby enhancing flow-field prediction accuracy\cite{hu2022flow}. 
The study showed that adding a radial-basis-function discriminator to a GAN improves flow-field reconstruction accuracy and stabilizes adversarial training — motivating our reuse of the same RBF-D as IG-GAN's discriminator, isolating the generator as the variable under study.

Transformer-based \cite{vaswani2017attention} generative models have also been introduced to aerodynamic data modeling due to their strong capability in capturing long-range dependencies. Jiang et al. proposed a Transformer-based decoder (TransCFD) to learn the mapping between aerodynamic geometries and flow fields, demonstrating accurate and efficient flow prediction while significantly reducing computational cost \cite{jiang2023transcfd}. 
In particular, Xu et al. proposed a self-supervised learning-based Transformer (SSL-Transformer) for flow reconstruction and prediction, which leverages large amounts of unlabeled data to improve model generalization and robustness under limited training samples \cite{xu2024self}. 
Moreover, attention-based and spatiotemporal transformer architectures have demonstrated the ability to capture complex aerodynamic interactions, enabling improved generalization across varying flow conditions and geometries.
Nevertheless, transformer models typically require large-scale training data and high computational resources, and their performance may degrade when training data is limited or noisy.

\begin{figure}[!t]
  \centering
  \includegraphics[width=\textwidth]{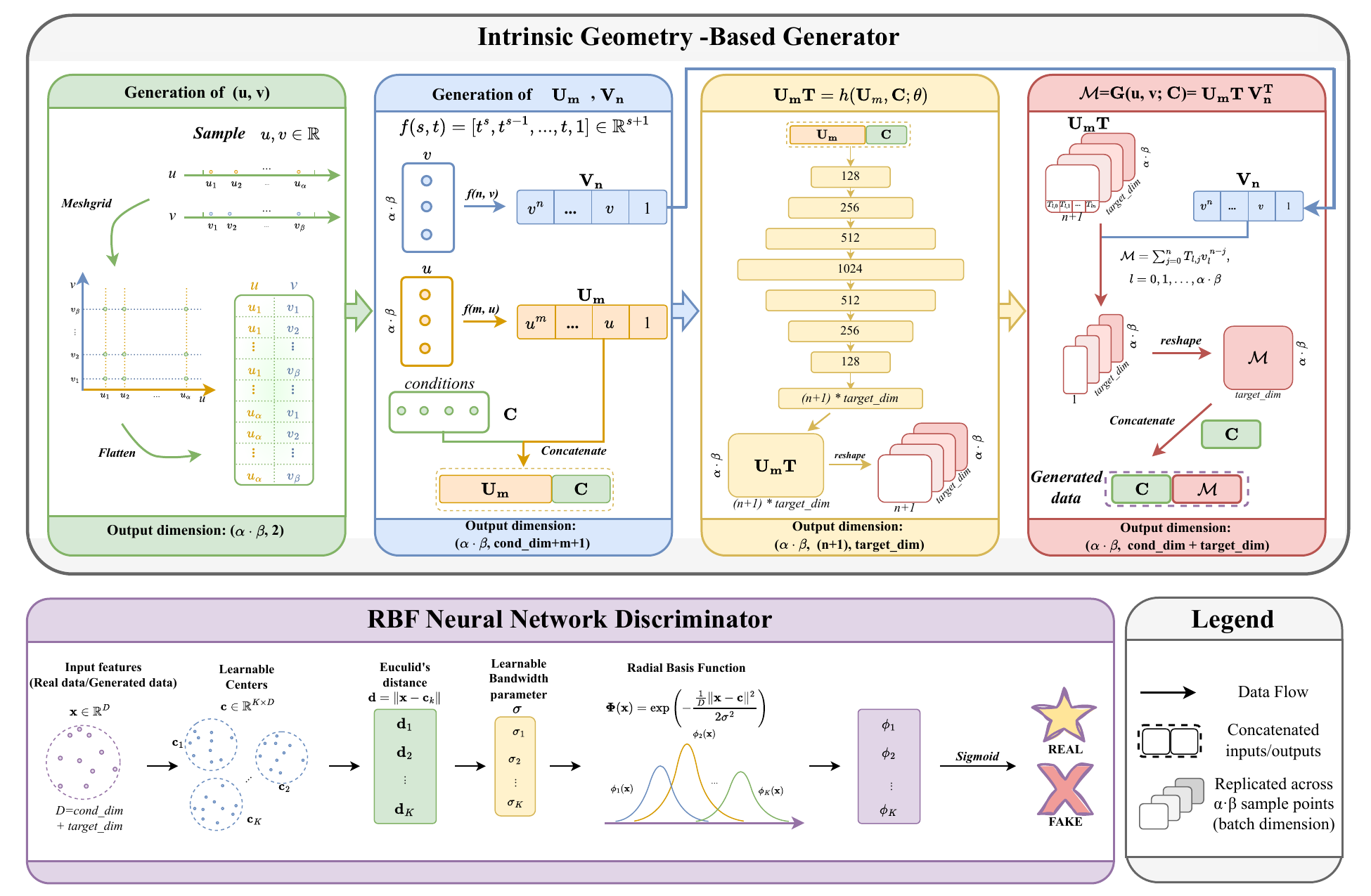}
  \caption{Overall architecture of the proposed IG-GAN model, including the Bézier-based generator module and the RBF neural-network discriminator module.}
  \label{fig:model_architecture}
\end{figure}

\section{METHOD}
\label{section_method}

\begin{table}[!t]
  \centering
  \caption{Notation used in Figure \ref{fig:model_architecture}.}
  \label{tab:notation}
  \small
  \begin{tabular}{ll}
  \toprule
  Symbol & Description \\
  \midrule
  \multicolumn{2}{l}{\textbf{Module 1: Generation of $(u,v)$}} \\
  \midrule
  $u, v$ & Sampled latent parametric coordinates \\
  $\alpha, \beta$ & Number of sampled points along the $u$- and $v$-directions \\
  $\alpha \cdot \beta$ & Total number of flattened grid points (output dimension of this module) \\
  \midrule
  \multicolumn{2}{l}{\textbf{Module 2: Generation of $\mathbf{U}_m$, $\mathbf{V}_n$}} \\
  \midrule
  $f(s,t)$ & Power-basis expansion function, $f(s,t) = [t^s, t^{s-1}, \dots, t, 1] \in \mathbb{R}^{s+1}$ \\
  $m, n$ & Polynomial degrees of the B\'ezier surface in $u$ and $v$ (Eq.~\eqref{equ_bezier}) \\
  $\mathbf{U}_m$ & Power-basis feature vector of $u$, $\mathbf{U}_m = f(m,u)$ \\
  $\mathbf{V}_n$ & Power-basis feature vector of $v$, $\mathbf{V}_n = f(n,v)$ \\
  $\mathbf{C}$ & Physical condition vector, $\mathbf{C} \in \mathbb{R}^{\text{cond\_dim}}$ \\
  \text{cond\_dim} & Dimensionality of the condition vector $\mathbf{C}$ \\
  \midrule
  \multicolumn{2}{l}{\textbf{Module 3: $\mathbf{U}_m\mathbf{T} = h(\mathbf{U}_m,\mathbf{C};\theta)$}} \\
  \midrule
  $h(\cdot;\theta)$ & Neural network predicting the coefficient representation from $(\mathbf{U}_m, \mathbf{C})$ \\
  $\theta$ & Trainable parameters of the generator network \\
  $\mathbf{U}_m\mathbf{T}$ & Network output; per-point partial coefficient tensor \\
  \text{target\_dim} & Dimensionality of the reconstructed physical quantity at each point \\
  \midrule
  \multicolumn{2}{l}{\textbf{Module 4: $\mathcal{M}=G(u,v;\mathbf{C})=\mathbf{U}_m\mathbf{T}\mathbf{V}_n^T$}} \\
  \midrule
  $G(u,v;\mathbf{C})$ & Full generator mapping (network $+$ analytic reconstruction) \\
  $T_{i,j}$ & Entries of the coefficient matrix $\mathbf{T}$ \\
  $l$ & Index over sampled grid points, $l = 0,1,\dots,\alpha\cdot\beta$ \\
  $\mathcal{M}$ & Reconstructed flow-field manifold, $\mathcal{M} \in \mathbb{R}^{\text{target\_dim}}$ per point \\
  $[\mathbf{C}, \mathcal{M}]$ & Generated data; concatenation of condition vector and reconstructed manifold \\
  \midrule
  \multicolumn{2}{l}{\textbf{Module 5: RBF Neural Network Discriminator}} \\
  \midrule
  $\mathbf{x}$ & Discriminator input, $\mathbf{x} = [\mathbf{C},\mathcal{M}] \in \mathbb{R}^D$ (real or generated) \\
  $D$ & Input dimensionality, $D = \text{cond\_dim} + \text{target\_dim}$ \\
  $\mathbf{c}_k$ & $k$-th learnable RBF center, $\mathbf{c}_k \in \mathbb{R}^D$ \\
  $K$ & Total number of RBF centers \\
  $\mathbf{d}_k$ & Euclidean distance between $\mathbf{x}$ and center $\mathbf{c}_k$, $\mathbf{d}_k = \|\mathbf{x}-\mathbf{c}_k\|$ \\
  $\sigma_k$ & Learnable bandwidth parameter for center $k$ \\
  $\Phi(\mathbf{x}), \phi_k(\mathbf{x})$ & Radial basis activation, $\phi_k(\mathbf{x}) = \exp\left(-\dfrac{1}{D}\dfrac{\|\mathbf{x}-\mathbf{c}_k\|^2}{2\sigma_k^2}\right)$ \\
  \bottomrule
  \end{tabular}
  \end{table}

Figure \ref{fig:model_architecture} illustrates the overall architecture of the proposed IG-GAN model, which consists of two major components: a generator built on an intrinsic geometric (B\'ezier-based) parameterization, and a discriminator implemented as a radial basis function (RBF) neural network.
Table \ref{tab:notation} summarizes the symbols used in Figure \ref{fig:model_architecture}.
In the generator, a grid of latent parametric coordinates $(u,v)$ is first sampled randomly and flattened over the two-dimensional parameter domain. 
Each coordinate is then expanded into a power-basis feature vector of degree $m$ (for $u$) or degree $n$ (for $v$), denoted $\mathbf{U}_m$ and $\mathbf{V}_n$, following equation \eqref{equ_bezier}. 
Only $\mathbf{U}_m$ is concatenated with the physical condition vector $\mathbf{C}$; this combined input is passed through a multilayer neural network that predicts, for every sampled point, the partial coefficient tensor $\mathbf{U}_m\mathbf{T}$ (equation \eqref{equ_coeff_prediction}). 
The basis vector $\mathbf{V}_n$ does not pass through the network---it is combined analytically with $\mathbf{U}_m\mathbf{T}$ via matrix multiplication to reconstruct the flow-field manifold $\mathcal{M}$. 
The reconstructed $\mathcal{M}$ is finally concatenated with $\mathbf{C}$ to form the generated aerodynamic sample $[\mathbf{C}, \mathcal{M}]$.
The discriminator receives real or generated samples $\mathbf{x} = [\mathbf{C}, \mathcal{M}] \in \mathbb{R}^D$, where $D = \text{cond\_dim} + \text{target\_dim}$, and evaluates their authenticity via an RBF network (RBF-D). 
The RBF-D compares $\mathbf{x}$ against $K$ learnable centers $\mathbf{c}_k \in \mathbb{R}^D$ using Euclidean distance, applies a learnable per-center bandwidth $\sigma_k$ to produce activations $\phi_k(\mathbf{x})$, and aggregates these through a final layer with sigmoid activation to classify the sample as real or fake.

The following two subsections describe the proposed framework in detail. We first define the Bézier-surface manifold construction, then describe how the generator learns it.

\subsection{Manifold Construction}

This section uses objects in 3-D space as an example. Given a set of control points:
\begin{equation}
\nonumber
\label{equ_dataset_matrix}
D=
\begin{bmatrix}
  P_{00} & P_{01} &\cdots & P_{0n} \\
  P_{10} & P_{ij} &\cdots & P_{1n} \\
  \vdots & \vdots &\ddots & \vdots \\  
  P_{m0} & P_{m1} &\cdots & P_{mn} 
\end{bmatrix}
\end{equation} 
where $P_{ij}=(x_{ij},y_{ij},z_{ij})$ represents the coordinates in the 3-D space for $i=0,1,2,\cdots,m$ and $j=0,1,2,\cdots,n$. 
The Bézier surface is then generated by the following formulation:

\begin{equation}
\label{equ_bezier}
\begin{cases}
  F(u, v)=\sum_{i=0}^m \sum_{j=0}^n P_{ij} B_{i, m}(u) B_{j, n}(v) \\
  B_{i, m}(u)=\frac{m !}{i !(m-i) !} u^{i}(1-u)^{m-i}\\
  B_{j, n}(v)=\frac{n !}{j !(n-j) !} v^{j}(1-v)^{n-j}
\end{cases}
\end{equation}
where $F(u,v)$ is the Bézier surface function. 
The parameters $u$ and $v$ $(u,v \in [0,1])$ define the 2-D latent parametric space, $m$ and $n$ denote the polynomial degrees of the surface, and $B_{i, m}(u)$ and $B_{j,n}(v)$ are the Bernstein basis coefficients.

The function $F(u, v): \mathbb{R}^2 \rightarrow \mathbb{R}^3$ serves as a geometric bridge, mapping the normalized 2-D parametric space $\{u,v\}$ to the 3-D space $\{x,y,z\}$. 
Consequently, the input to $F(u,v)$ is a 2-D parametric vector $\left(u,v\right)$ and the output is a 3-D state vector $\left(x,y,z\right)$.

The Jacobian matrix of $F(u,v)$ is:
\begin{equation}
\label{equ_jacobin_F}
J_F=d(F(u,v))=
\begin{bmatrix}
  \frac{\partial F^x}{\partial u} & \frac{\partial F^x}{\partial v} \\[6pt]
  \frac{\partial F^y}{\partial u} & \frac{\partial F^y}{\partial v} \\[6pt]
  \frac{\partial F^z}{\partial u} & \frac{\partial F^z}{\partial v}
\end{bmatrix}.
\end{equation} 
where $F^x$, $F^y$, and $F^z$ denote the output by the mapping $F(u,v)$. 
The terms $\frac{\partial F^x}{\partial u}$ and $\frac{\partial F^x}{\partial v}$ represent the partial derivatives of the $x$-component with respect to the latent parameters $u$ and $v$, respectively.

In this framework, the rank of $J_F$ is maintained at 2, ensuring that $F(u,v)$ constitutes a smooth immersion. 
If the function $F(u,v)$ has no self-intersections, then $F(u,v)$ is a smooth embedding in $\mathbb{R}^3$. 
To represent complex, global flow fields, these individual embeddings are concatenated to form a unified piecewise smooth manifold:
\begin{equation}
\label{equ_manifold_union}
M(D;u,v)=\mathrm{U}_{q=1}^{\mathcal{Q}} F_q(u, v)
\end{equation} 
where $M(D;u,v)$ (denoted as $\mathcal{M}$) is the complete reconstructed manifold, $F_q(u,v)$ represents the $q$-th smooth embedding, and $\mathcal{Q}$ is the total number of segments. 
Through this manifold construction, the intricate geometric features of the 3-D flow field are efficiently parameterized and reconstructed within a continuous 2-D space.

\subsection{Intrinsic Geometry-Based Generator}

To establish a general framework for flow field reconstruction, we define the constructed Bézier surface $\mathcal{M}$ using arbitrary degrees $m$ and $n$. 
This surface acts as a geometric parameterization formed by the tensor product of Bernstein basis functions.
In equation \eqref{equ_bezier}, we let 
\begin{equation}
\begin{cases}
  \mathbf{B}_m(u) = [B_{0,m}(u), \dots, B_{i, m}(u), \dots, B_{m,m}(u)] \\
  \mathbf{B}_n(v) = [B_{0,n}(v), \dots, B_{j, n}(v), \dots, B_{n,n}(v)]
\end{cases}
\end{equation}
the surface is then expressed as:
\begin{equation}
\label{Bézier_surface}
\mathcal{M} = F(u,v) = \mathbf{B}_m(u) \, \mathbf{P} \, \mathbf{B}_n(v)^T.
\end{equation}

To further refine this, we expand the Bernstein basis function $B_{i, n}(t)$ into its power basis form.
By applying the binomial theorem to $(1-t)^{n-i}$ and collecting coefficients for each power of $t$, we rewrite the basis function as a polynomial $B_{i, n}(t) = a_{i,0} t^n + a_{i,1} t^{n-1} + \dots + a_{i,n}$. 
This allows the basis vector $\mathbf{B}_m(u)$ to be expressed as the product of the power basis vector $\mathbf{U}_m$ and a structured transformation matrix $\mathbf{A}_m$:
\begin{equation}
\begin{aligned}
\mathbf{B}_m(u) &= \begin{bmatrix} u^m & u^{m-1} & \dots & 1 \end{bmatrix}
&\times 
\begin{bmatrix}
a_{0,0} & a_{1,0} & \dots & a_{m,0} \\
a_{0,1} & a_{1,1} & \dots & a_{m,1} \\
\vdots & \vdots & \ddots & \vdots \\
a_{0,m} & a_{1,m} & \dots & a_{m,m}
\end{bmatrix} = \mathbf{U}_m \mathbf{A}_m.
\end{aligned}
\end{equation}

Following the same logic for $v$, we define $\mathbf{B}_n(v) = \mathbf{V}_n \mathbf{A}_n$. 
Substituting these into the equation \eqref{Bézier_surface}:
\begin{equation}
\mathcal{M} = \mathbf{U}_m \mathbf{A}_m \mathbf{P} (\mathbf{V}_n \mathbf{A}_n)^T = \mathbf{U}_m \mathbf{A}_m \mathbf{P} \mathbf{A}_n^T \mathbf{V}_n^T.
\end{equation}

We define the generator $G(u,v;\mathbf{C})$ as the complete mapping from the latent parametric coordinates and physical conditions to the reconstructed aerodynamic sample. 
This mapping consists of two stages: a learned stage, in which a neural network predicts a coefficient matrix from the sampled input, and an analytic stage, in which this coefficient matrix is combined with the Bézier basis vectors to reconstruct the manifold.

Let $h(\cdot;\theta)$ denote the neural network with trainable parameters $\theta$. 
Given the vector $\mathbf{U}_m$ and the condition vector $\mathbf{C}$, the network predicts the coefficient matrix $\mathbf{T}$:
\begin{equation}
\mathbf{T} = \mathbf{A}_m \mathbf{P} \mathbf{A}_n^T,
\end{equation}
\begin{equation}
  \label{equ_coeff_prediction}
  h(\mathbf{U}_m, \mathbf{C}; \theta) = \mathbf{U}_m \mathbf{T},
\end{equation}
where $\mathbf{A}_m \mathbf{P} \mathbf{A}_n^T$ is the closed-form coefficient matrix defined by the Bézier control points $\mathbf{P}$ and the transformation matrices $\mathbf{A}_m$, $\mathbf{A}_n$ introduced in equation \eqref{Bézier_surface}. 
Rather than supervising $\mathbf{T}$ directly against this closed form, the network learns to produce $\mathbf{T}$ end-to-end through the reconstruction objective described below; equation \eqref{equ_coeff_prediction} is included to show the correspondence between the learned representation and the underlying Bézier coefficient structure.
Only $\mathbf{U}_m$ and $\mathbf{C}$ are passed into the network; the basis vector $\mathbf{V}_n$ is not used at this stage and instead enters only in the subsequent analytic reconstruction.

The full generator is then defined as:
\begin{equation}
\label{equ_full_generator}
\mathcal{M} = G(u,v;\mathbf{C}) = h(\mathbf{U}_m, \mathbf{C};\theta) \, \mathbf{V}_n^T = \mathbf{U}_m \mathbf{T} \mathbf{V}_n^T,
\end{equation}
where $G(u,v;\mathbf{C})$ denotes the complete generative process, combining the learned coefficient matrix $\mathbf{T}$ with the analytic Bézier basis vectors $\mathbf{U}_m$ and $\mathbf{V}_n$ to reconstruct the flow-field manifold $\mathcal{M}$.

This formulation corresponds to the Intrinsic Geometry-Based Generator module in Figure \ref{fig:model_architecture}, where $\mathbf{U}_m$ and $\mathbf{V}_n$ provide the intrinsic geometric inputs, the neural network $h(\cdot;\theta)$ predicts the coefficient matrix $\mathbf{U}_m \mathbf{T}$, and the matrix product $\mathbf{U}_m \mathbf{T} \mathbf{V}_n^T$ reconstructs the generated aerodynamic data.

\section{EXPERIMENTS AND RESULTS ANALYSIS}
\label{section_experiments_and_results_analysis}

We conduct two experiments on the Burgers' and ONERA M6 datasets to validate IG-GAN against three baselines (RBF-GAN\cite{hu2022flow}, Riemannian geometric features-incorporated learning (RGFiL) method\cite{hu2025learning}, and SSL-Transformer\cite{xu2024self}) on flow-field reconstruction and aerodynamic coefficient prediction, respectively.

\subsection{Experiment I: Nonlinear Parametric Flow-Field Reconstruction on the Burgers' Dataset}

\subsubsection{Burgers' dataset and preprocessing}

\begin{figure}[ht]
  \centering
  \includegraphics[width=0.55\columnwidth]{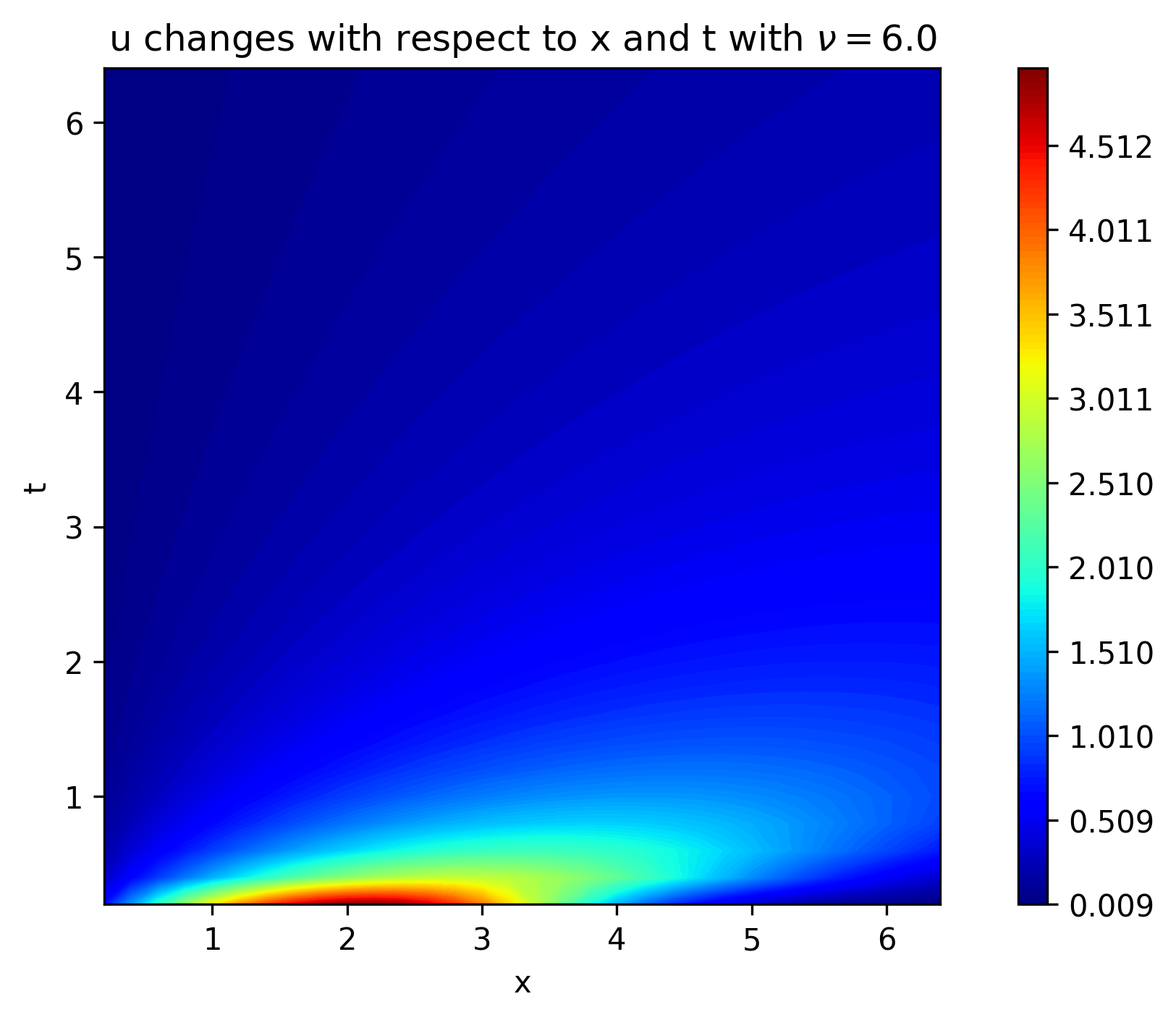}
  \caption{Ground-truth solution $u$ of the Burgers' equation with $\nu=6.0$ as a function of displacement $x$ and time $t$.}
  \label{fig:true_nu6_0_heatmap}
\end{figure}

Burgers' equation is a 1-D Partial differential equation (PDE) that expresses the movement of a shockwave across a tube
\begin{equation}
\frac{\partial u}{\partial t} + u \frac{\partial u}{\partial x} = v \frac{\partial^2 u}{\partial x^2}
\end{equation}
where $u$ denotes the velocity of the fluid (the shockwave), $t$ denotes the time, $x$ denotes the displacement, and $\nu$ denotes the viscosity coefficient. 
The PDE constrains the solution space through nonlinear dynamics and diffusion. 
Consequently, physically valid solutions exhibit strong spatial-temporal correlations and occupy only a low-dimensional nonlinear manifold embedded in the high-dimensional observation space. 
This characteristic makes the Burgers' dataset suitable for evaluating the ability of generative models to capture intrinsic manifold structures and produce physically consistent flow fields.
The variation range of the flow parameters ($t$, $x$, and $\nu$) is from 0.2 to 6.4, and the step is 0.2. 
Consequently, this dataset contains 32,768 samples. 
Figure \ref{fig:true_nu6_0_heatmap} shows the variation in velocity $u$ with respect to time $t$ and displacement $x$ under the condition $\nu$ = 6.0. 
In this figure, the red part in the lower left corner describes the variations in high-velocity flows. 
Therefore, Burgers' dataset contains limited high-velocity nonlinear data that can be used to verify whether the model can generate high-velocity data accurately.

\begin{figure}[!t]
  \centering
  \includegraphics[width=\textwidth]{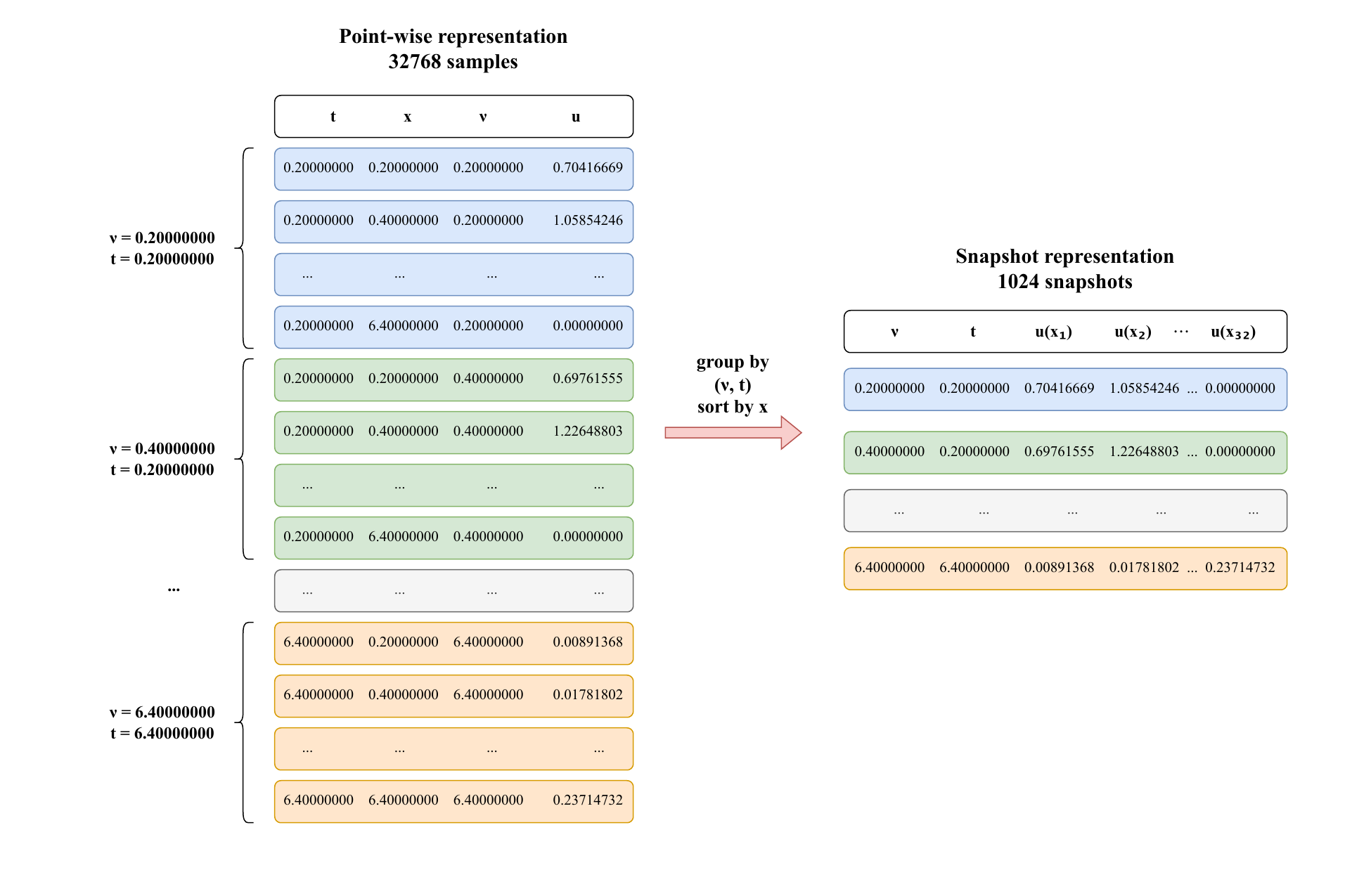}
  \caption{Restructuring of the Burgers' dataset from point-wise samples into flow-field snapshots. Rows sharing the same $(\nu, t)$ pair are grouped and sorted by spatial coordinate $x$ to form a one-dimensional snapshot $u(x;\nu,t)$ of 32 spatial values. The full dataset of 32,768 point-wise samples reduces to 1,024 snapshots.}
  \label{fig:burgers_snapshot}
\end{figure}

For IG-GAN, RBF-GAN, and RGFiL, each sample was represented as a single data point $(t,x,\nu,u)$, where $(t,x,\nu)$ were used as the physical condition variables and $u$ was treated as the target response. 
The full dataset was randomly divided into training, validation, and test sets with a ratio of 8:1:1. 

For the SSL-Transformer model, the original point-wise data were reorganized into snapshot form, as shown in the Figure \ref{fig:burgers_snapshot} .
Specifically, for each pair of viscosity coefficient $\nu$ and time $t$, the corresponding velocity values along the spatial coordinate $x$ were collected and sorted to form a complete one-dimensional flow-field snapshot $u(x;\nu,t)$. 
In this way, the Burgers' dataset was transformed from point-wise samples into field-level samples, allowing the SSL-Transformer to perform flow-field reconstruction based on partial observations. 
The generated snapshots were then split into training, validation, and test sets using the same 8:1:1 ratio.

\begin{table}[!t]
  \caption{Input Representation and split settings for the Burgers' dataset.}
  \label{tab:burgers_preprocess_settings}
  \centering
  \footnotesize
  \renewcommand{\arraystretch}{1.1}
  \begin{tabular*}{\textwidth}{@{\extracolsep{\fill}}ccc@{}}
  \toprule
  Model & Input Representation & Train/Val/Test \\
  \midrule
  IG-GAN & Point $(t,x,\nu,u)$ & 26,214/3,277/3,277 \\
  RBF-GAN & Point $(t,x,\nu,u)$ & 26,214/3,277/3,277 \\
  RGFiL & Point $(t,x,\nu)\!\to\!u$ & 26,214/3,277/3,277 \\
  SSL-Transformer & Snapshot $u(x;\nu,t)$ & 819/102/103 \\
  \bottomrule
  \end{tabular*}
  \end{table}

Table~\ref{tab:burgers_preprocess_settings} summarizes the input representation and split settings used for the four compared models. The GAN-based and RGFiL models use point-wise samples, whereas the SSL-Transformer uses snapshot-level samples. All input and output variables were normalized using Min-Max normalization. During evaluation, the predicted results were transformed back to the original physical scale before calculating the final error metrics.

\subsubsection{Model settings}

For all models, the best checkpoint is selected according to the minimum validation MSE and is then evaluated on the test set.

\begin{table}[!t]
  \caption{Network configurations of the compared models.}
  \label{tab:burgers_architecture}
  \centering
  \footnotesize
  \renewcommand{\arraystretch}{1.35}
  \setlength{\tabcolsep}{6pt}
  
  \begin{tabularx}{\textwidth}{@{}C{2.5cm}C{3.3cm}Y@{}}
  \toprule
  Model &
  Component &
  Configuration \\
  \midrule
  
  \multirow{2}{*}{IG-GAN}
  &
  \makecell[c]{Generator\\(IG-G)}
  &
  Hidden layers: 128, 256, 512, 1024, 512, 256, 128;
  LeakyReLU activation.
  \\[6pt]
  
  &
  \makecell[c]{Discriminator\\(RBF-D)}
  &
  1024 RBF centers;
  Sigmoid output.
  \\
  
  \midrule
  
  \multirow{2}{*}{RBF-GAN}
  &
  \makecell[c]{Generator\\(FCN)}
  &
  Hidden layers: 128, 256, 512, 1024, 512, 256, 128;
  LeakyReLU activation.
  \\[6pt]
  
  &
  \makecell[c]{Discriminator\\(RBF-D)}
  &
  1024 RBF centers;
  Sigmoid output.
  \\
  
  \midrule
  
  \multirow{2}{*}{RGFiL}
  &
  \makecell[c]{Expert network\\(FCN, $\times3$)}
  &
  Hidden layers: 64, 128, 256, 128, 64;
  LeakyReLU activation.
  \\[6pt]
  
  &
  \makecell[c]{Gating network\\(FCN)}
  &
  Hidden layers: 64, 128, 256, 512, 1024, 512, 256, 128, 64;
  Softmax output.
  \\
  
  \midrule
  
  \multirow{2}{*}{SSL-Transformer}
  &
  Encoder
  &
  Embedding dimension: 128;
  Four Transformer layers;
  One attention head;
  Latent dimension: 128;
  Temporal window $tw=1$.
  \\[6pt]
  
  &
  Decoder
  &
  Fourier projection;
  Cross-attention;
  Linear attention;
  Scale factor: 2.0.
  \\
  
  \bottomrule
  \end{tabularx}
  \end{table}

\begin{table}[!t]
  \caption{Input/output dimensions of the four compared models in Experiment I.}
  \label{tab:exp1_io}
  \centering
  \footnotesize
  
  \begin{tabularx}{\textwidth}{@{}C{2.4cm}C{3.4cm}Y@{}}
  \toprule
  Model & Component & Input / Output \\
  \midrule
  \multirow{2}{*}{IG-GAN}
            & Generator
            & $7(3+4)\rightarrow4$ \\

            & Discriminator
            & $4\rightarrow1$ \\

  \midrule
  
  \multirow{2}{*}{RBF-GAN}
            & Generator
            & $67(3+64)\rightarrow4$ \\

            & Discriminator
            & $4\rightarrow1$ \\

  \midrule

  \multirow{2}{*}{RGFiL}
            & Expert network
            & $3 \rightarrow1 $ \\

            & Gating network
            & $3\rightarrow 3$ \\

  \midrule
  
  SSL-Transformer & Encoder--Decoder & Snapshot ($32$) $\rightarrow$ Snapshot ($32$) \\

  \bottomrule
  \end{tabularx}
  
  \end{table}

The proposed IG-GAN is formulated as a conditional adversarial model. 
The Intrinsic geometry-based generator(IG-G) concatenates the physical condition variables $(t,x,\nu)$ and $(u^3, u^2, u, 1)$ as inputs, passes the result through a FCN, the full dimensional specification is given in Table~\ref{tab:burgers_architecture}, the resulting coefficient vector performs an operation on $(v^3, v^2, v, 1)$ and obtains the scalar prediction $\hat{u}$.
The discriminator used in IG-GAN is an RBF neural network operating in the four-dimensional state space $(t,x,\nu,u)$\cite{hu2022flow}. 
It contains 1,024 trainable RBF centers followed by a fully connected output layer with sigmoid activation. 

RBF-GAN adopts the same RBF discriminator structure as IG-GAN, and uses a fully connected network(FCN) as the generator. 
The generator receives the condition variables $(t,x,\nu)$ together with a 64-dimensional random noise vector, giving an input dimension of $3+64=67$, and outputs the predicted scalar velocity $\hat{u}$. 
The hidden layer dimensions and activation functions are listed in Table~\ref{tab:burgers_architecture}. 
The difference between IG-GAN and RBF-GAN is that IG-GAN uses an intrinsic-geometry generator, while RBF-GAN uses a FCN.

The RGFiL baseline is implemented as a multi-branch fully connected model. 
It contains three expert subnetworks and a context-gating network. 
Each expert takes $(t,x,\nu)$ as input and predicts a scalar velocity response, while the gating network estimates the adaptive weights assigned to the three experts through a softmax operation. 
The input and output dimensions of each component, together with their hidden layer configurations, are given in Table~\ref{tab:burgers_architecture}. LeakyReLU activations are used throughout.

The SSL-Transformer baseline is configured as an encoder-decoder reconstruction model for snapshot-level Burgers flow fields. 
Each snapshot contains 32 spatial points, and the position coordinates are represented by $(x,\nu)$. 
The encoder uses an irregular spatio-temporal transformer and the decoder reconstructs the complete flow field from sparse observations through Fourier coordinate projection, cross-attention, and linear attention. 
The detailed architectural parameters are listed in Table~\ref{tab:burgers_architecture}. 
During training, sparse observation is imposed by retaining a random subset of spatial points, with the observation ratio set to 0.2.

The input and output dimensions of each model are shown in the Table~\ref{tab:exp1_io}.
The shared training settings are as follows. 
All models are trained for 2500 epochs with learning rate $1\times10^{-4}$, and the final test evaluation uses the checkpoint with the minimum validation MSE. 
The Adam optimizers use $\beta_1=0.5$ and $\beta_2=0.999$. 

The GAN-based models are trained with binary cross-entropy loss for adversarial optimization, while RGFiL and SSL-Transformer are trained as direct regression or reconstruction baselines. 
This setup allows the comparison to cover both adversarial generation-based reconstruction and non-adversarial predictive modeling.

\subsubsection{Results and Analysis}

The reconstruction performance of the four compared models on the test set is summarized in Table~\ref{tab:burgers_test_performance}. 

\begin{table}[!t]
\caption{Test-set performance of the four compared models on the Burgers' dataset.}
\label{tab:burgers_test_performance}
\centering
\setlength{\tabcolsep}{4pt}
\footnotesize
\renewcommand{\arraystretch}{1.15}
\begin{tabular*}{\textwidth}{@{\extracolsep{\fill}}cccccc@{}}
\toprule
Model & MAE & MSE & RMSE & $R^2$ & Corr. \\
\midrule
IG-GAN & $\mathbf{4.345\times10^{-3}}$ & $\mathbf{1.004\times10^{-4}}$ & $\mathbf{1.002\times10^{-2}}$ & $\mathbf{0.9995}$ & $\mathbf{0.9998}$ \\
RBF-GAN & $5.799\times10^{-3}$ & $2.059\times10^{-4}$ & $1.435\times10^{-2}$ & 0.9990 & 0.9995 \\
RGFiL & $3.016\times10^{-2}$ & $7.303\times10^{-3}$ & $8.545\times10^{-2}$ & 0.9659 & 0.9840 \\
SSL-Transformer & $2.829\times10^{-2}$ & $3.871\times10^{-3}$ & $6.222\times10^{-2}$ & 0.9884 & 0.9942 \\
\bottomrule
\end{tabular*}
\end{table}

IG-GAN obtains the lowest error among all compared models, with an MAE of $4.345\times10^{-3}$, an MSE of $1.004\times10^{-4}$, and an RMSE of $1.002\times10^{-2}$. 
It also achieves the highest $R^2$ and correlation coefficient, indicating that the predicted velocity values are highly consistent with the ground-truth solution. 
Compared with RBF-GAN, IG-GAN reduces MAE by approximately $25.07\%$, MSE by $51.19\%$, and RMSE by $30.17\%$. 
Since these two models share the same RBF discriminator and mainly differ in the generator structure, this improvement suggests that the intrinsic-geometry generator provides a more effective representation for the nonlinear Burgers' flow-field reconstruction task.

The non-adversarial baselines show larger errors. 
RGFiL produces the highest MSE and the lowest $R^2$, indicating that the mixture-of-experts regression structure is less effective in resolving the nonlinear flow response over the full parameter range. 
SSL-Transformer performs better than RGFiL, but its error remains much higher than that of the two GAN-based models. 
This may be related to the snapshot-level reconstruction formulation, where the model reconstructs the full field from sparse observations rather than directly learning the point-wise mapping from $(t,x,\nu)$ to $u$.

\begin{figure}[!t]
  \centering
  \includegraphics[width=\textwidth]{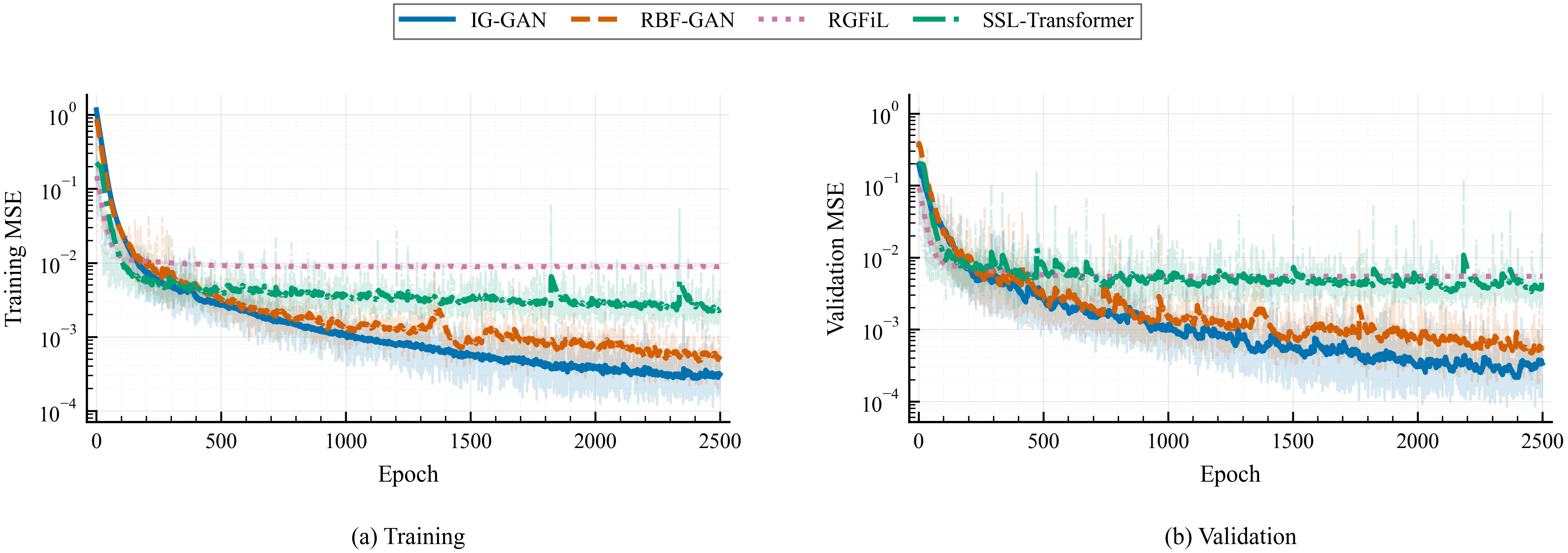}
  \caption{Training and validation MSE curves of the four compared models on the Burgers' dataset.}
  \label{fig:burgers_mse_curves}
\end{figure}

The most notable feature of Figure~\ref{fig:burgers_mse_curves} is RGFiL's training curve, which plateaus near $10^{-2}$ MSE almost immediately and shows essentially no further improvement over 2,500 epochs — consistent with its highest test-set MSE in Table~\ref{tab:burgers_test_performance} and suggesting the mixture-of-experts structure saturates early on this task. 
IG-GAN and RBF-GAN both continue decreasing well past epoch 1,000, with IG-GAN's validation curve consistently below RBF-GAN's after roughly epoch 300, matching the final ranking in Table~\ref{tab:burgers_test_performance}. SSL-Transformer's validation curve is visibly noisier throughout, which may reflect its much smaller effective sample count.

\begin{figure}[!t]
  \centering
  \includegraphics[width=\textwidth]{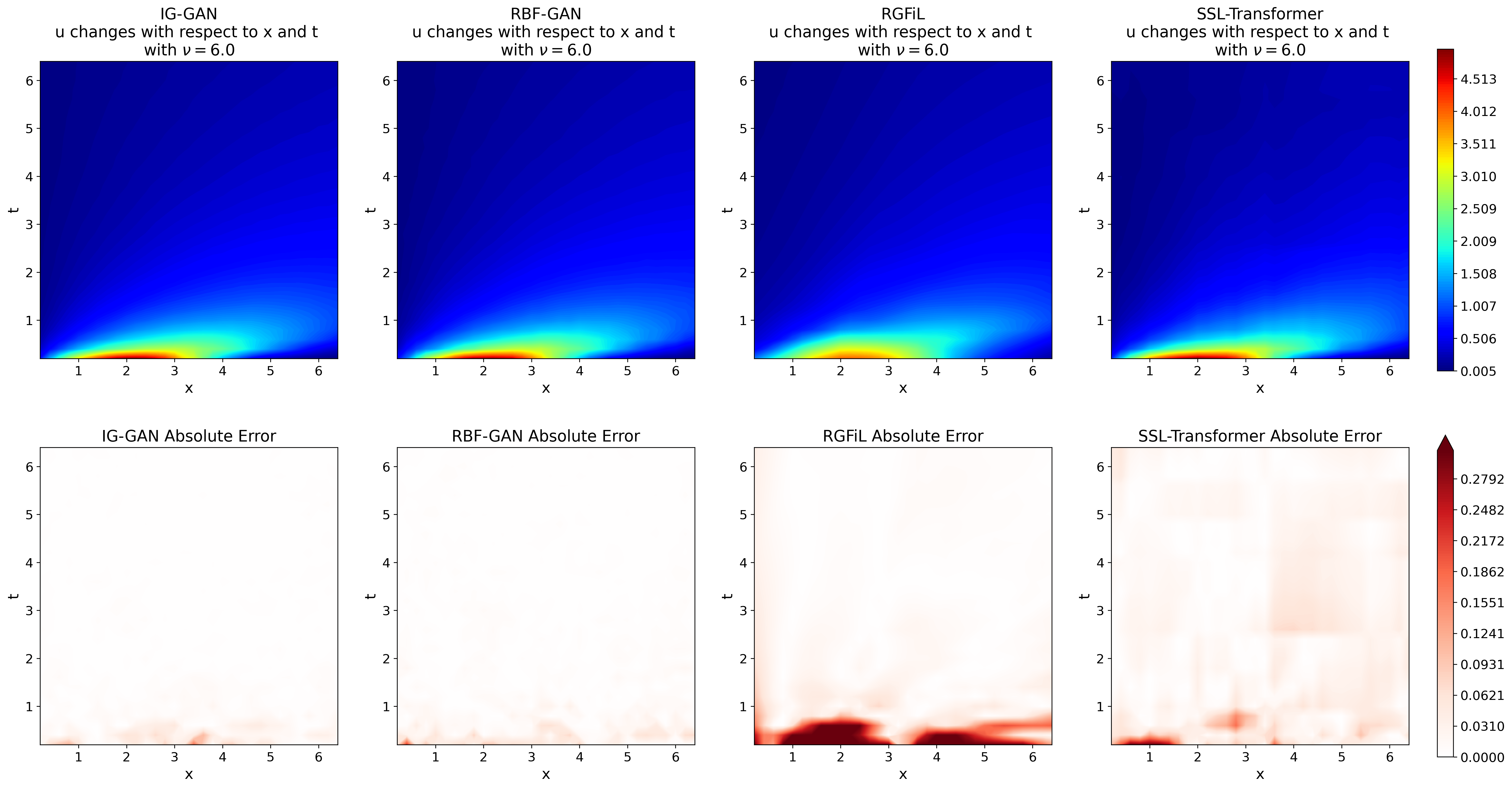}
  \caption{Reconstructed flow fields and absolute-error distributions of the four compared models at $\nu=6.0$.}
  \label{fig:burgers_nu6_0_reconstruction}
\end{figure}

\begin{figure}[!t]
  \centering
  \includegraphics[width=\textwidth]{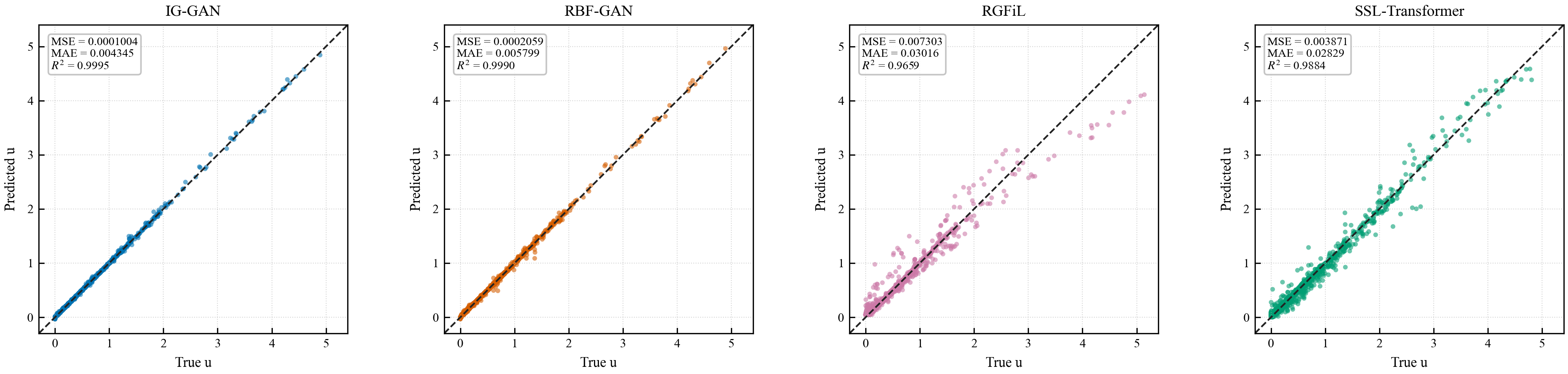}
  \caption{Predicted versus true velocity $u$ for the four compared models on the held-out Burgers' test set. The dashed line denotes the ideal 45$^\circ$ reference.}
  \label{fig:burgers_45degree}
\end{figure}

Figure~\ref{fig:burgers_nu6_0_reconstruction} shows the reconstructed field at $\nu=6.0$. 
IG-GAN's error map (bottom row) is nearly uniform and close to zero across the domain, including the high-velocity shock region near t = 0.2–0.8 — there is no visible error concentration that would suggest mode collapse. 
RBF-GAN's errors are similarly low overall but noticeably higher in that same high-velocity band. 
RGFiL shows a distinct band of large error (dark red, Figure~\ref{fig:burgers_nu6_0_reconstruction} bottom row third panel) concentrated in exactly this shock region, indicating it struggles most where the solution is least linear. 
SSL-Transformer's errors are more diffuse and less spatially concentrated, but larger in magnitude than either GAN-based model's.

Figure~\ref{fig:burgers_45degree} further evaluates point-wise prediction fidelity through predicted-versus-true scatter plots. IG-GAN shows the tightest clustering around the ideal 45$^\circ$ reference line across the full velocity range, demonstrating that the model maintains accurate predictions in both low-velocity and high-velocity regions. RBF-GAN also follows the reference line closely, but the scatter is slightly wider, especially in the mid-to-high velocity range. The RGFiL scatter plot is more dispersed, indicating larger point-wise deviations and weaker control of nonlinear response variations. SSL-Transformer shows intermediate behavior: its predictions retain the overall trend but display more dispersion than the GAN-based models. These scatter plots confirm that IG-GAN achieves both low average error and strong point-wise agreement with the ground-truth Burgers' solution.

\subsection{Experiment II: Aerodynamic Coefficient Prediction on the ONERA M6 Dataset}

\subsubsection{ONERA M6 dataset and preprocessing}

The ONERA M6 dataset is used to evaluate the ability of the proposed model to predict aerodynamic coefficients under different flow states. 
We generated 770,000 grids on the surface and the far-field region of the ONERA M6 wing. 
And we calculated 755 samples by SU2\cite{economon2016su2} to describe the $C_P$ variation on the wing surface. 
Each sample contains three flow-condition variables, namely Mach number (Ma), angle of attack(AoA), and Reynolds number(Re), together with aerodynamic force and moment coefficients. 
The ONERA M6 dataset also exhibits an intrinsic manifold structure. 
Although each sample is represented by a high-dimensional pressure coefficient ($C_P$) distribution over the wing surface, the flow field is determined by a small number of physical parameters. 
Under the equations of compressible fluid dynamics, these parameters constrain the physically realizable pressure distributions to a low-dimensional nonlinear manifold embedded in the high-dimensional solution space. 
As the flow conditions vary continuously, the corresponding pressure distributions evolve smoothly along this manifold. 
Therefore, the ONERA M6 dataset provides an appropriate scene for evaluating whether a generative model can accurately learn the intrinsic manifold structure of aerodynamic flow fields.
The format and variation range of the input flow parameters are summarized in Table~\ref{tab:m6_flow_parameters}, and the aerodynamic coefficients are listed in Table~\ref{tab:m6_aero_outputs}.

\begin{table}[t]
  \caption{Input flow parameters of the ONERA M6 dataset.}
  \label{tab:m6_flow_parameters}
  \centering
  \footnotesize
  \setlength{\tabcolsep}{3pt}
  \renewcommand{\arraystretch}{1.2}
  \renewcommand{\tabularxcolumn}[1]{m{#1}}
  \begin{tabularx}{\columnwidth}{@{}C{1.15cm}YYY@{}}
  \toprule
  \makecell[c]{Item} & \makecell[c]{$Ma$\\(Mach number)} & \makecell[c]{AoA $(\alpha)$\\(Angle of attack)} & \makecell[c]{$Re$\\(Reynolds number)} \\
  \midrule
  Start & $0.8045$ & $1.06^\circ$ & $8.03\times10^6$ \\
  End & $0.8395$ & $4.81^\circ$ & $1.30\times10^7$ \\
  Step & $0.005$ & $0.25^\circ$ & $2.5\times10^5$ \\
  \bottomrule
  \end{tabularx}
\end{table}

\begin{table}[t]
  \caption{Data format of the aerodynamic characteristics in the ONERA M6 dataset.}
  \label{tab:m6_aero_outputs}
  \centering
  \footnotesize
  \setlength{\tabcolsep}{4pt}
  \renewcommand{\arraystretch}{1.12}
  \begin{tabularx}{\textwidth}{@{}C{2.2cm}YYY@{}}
  \toprule
  Variables & $C_L$ & $C_D$ & $C_{SF}$ \\
  Significance & coefficient of lift & coefficient of drag & coefficient of skin friction \\
  \midrule
  Variables & $C_{Mx}$ & $C_{My}$ & $C_{Mz}$ \\
  Significance & coefficient of rolling moment & coefficient of pitching moment & coefficient of yawing moment \\
  \midrule
  Variables & $C_{Fx}$ & $C_{Fy}$ & $C_{Fz}$ \\
  Significance & $X$ component coefficient of friction & $Y$ component coefficient of friction & $Z$ component coefficient of friction \\
  \bottomrule
  \end{tabularx}
  \end{table}

The dataset was randomly divided into training, validation, and test sets with an 8:1:1 ratio, resulting in 604 training samples, 75 validation samples, and 76 test samples. 
The input variables and output variables were normalized separately by Min-Max normalization. 
During evaluation, all predicted coefficients were transformed back to their original physical scale before calculating the final metrics.

For IG-GAN, RBF-GAN, and RGFiL, the three flow-condition variables of the dataset were used as inputs and the nine aerodynamic coefficients were used as targets. 
For SSL-Transformer, the same tabular sample was reorganized into a compact snapshot form. 
Specifically, the nine target coefficients were treated as output tokens, and each token was associated with the corresponding normalized flow-condition information and a target-position index. 
This representation allows SSL-Transformer to perform coefficient reconstruction in a token-based form while using the same training/validation/test split as the other models. 
Table~\ref{tab:m6_preprocess_settings} summarizes the data representation and split settings used in this experiment.

\begin{table}[t]
  \caption{Data representation and split settings for the ONERA M6 dataset.}
  \label{tab:m6_preprocess_settings}
  \centering
  \footnotesize
  \renewcommand{\arraystretch}{1.15}
  \setlength{\tabcolsep}{4pt}
  
  \begin{tabularx}{\columnwidth}{@{}C{1.8cm}YC{1.8cm}@{}}
  \toprule
  Model &
  Data Representation &
  Split \\
  \midrule
  IG-GAN &
  Point $(Ma,AoA,Re)\rightarrow 9$ coefficients &
  604/75/76 \\
  
  RBF-GAN &
  Point $(Ma,AoA,Re)\rightarrow 9$ coefficients &
  604/75/76 \\
  
  RGFiL &
  Point $(Ma,AoA,Re)\rightarrow 9$ coefficients &
  604/75/76 \\
  
  \makecell{SSL-\\Transformer} &
  Snapshot with 9 coefficient tokens &
  604/75/76 \\
  \bottomrule
  \end{tabularx}
\end{table}

\subsubsection{Model settings}

In this experiment, all models were trained with learning rate $1\times10^{-4}$, and the checkpoint with the minimum validation MSE was used for final testing. 
The Adam optimizers used $\beta_1=0.5$ and $\beta_2=0.999$. 

\begin{table}[!t]
  \caption{Input/output dimensions of the four compared models in Experiment II.}
  \label{tab:m6_architecture}
  \centering
  \footnotesize
  
  \begin{tabularx}{\textwidth}{@{}C{2.4cm}C{3.4cm}Y@{}}
  \toprule
  Model & Component & Input / Output \\
  \midrule
  \multirow{2}{*}{IG-GAN}
            & Generator
            & $7(3+4)\rightarrow12$ \\

            & Discriminator
            & $12\rightarrow1$ \\

  \midrule
  
  \multirow{2}{*}{RBF-GAN}
            & Generator
            & $67(3+64)\rightarrow12$ \\

            & Discriminator
            & $12\rightarrow1$ \\

  \midrule

  \multirow{2}{*}{RGFiL}
            & Expert network
            & $3 \rightarrow1 $ \\

            & Gating network
            & $3\rightarrow 9$ \\

  \midrule
  
  SSL-Transformer & Encoder--Decoder & Snapshot ($12$) $\rightarrow$ Snapshot ($12$) \\

  \bottomrule
  \end{tabularx}
  
  \end{table}

For IG-GAN, the generator takes the three normalized flow-condition variables $(Ma,AoA,Re)$ and $(u^3, u^2, u, 1)$ as inputs, and outputs the nine aerodynamic coefficients simultaneously. 
The input dimension of the fully connected generator is therefore $4+3=7$, and the output dimension becomes $4\times9=36$; the hidden layer configuration remains identical to that in Experiment I, as noted in Table~\ref{tab:burgers_architecture}. 
The generated coefficients are concatenated with the flow-condition variables before being sent to the discriminator. 
Accordingly, the RBF discriminator operates in a 12-dimensional state space consisting of three input variables and nine output coefficients, with 1,024 trainable RBF centers and a sigmoid output layer, as in Experiment I.

RBF-GAN uses the same 12-dimensional RBF discriminator as IG-GAN, but its generator is the conventional fully connected generator conditioned on $(Ma,AoA,Re)$ and a 64-dimensional random noise vector, giving an input dimension of $3+64=67$. 
The output layer is expanded to predict nine coefficients simultaneously; all hidden layer dimensions are unchanged from Experiment I. 
This allows the comparison between IG-GAN and RBF-GAN to focus on whether the intrinsic-geometry generator remains beneficial for a sparse aerodynamic-coefficient dataset.

For the RGFiL baseline, the scalar multi-branch structure used in Experiment I is extended to the multi-output M6 task by assigning one independent RGFiL module to each of the nine aerodynamic coefficients. 
Each module retains the same three-expert and softmax gating architecture as in Experiment~1 (Table~\ref{tab:burgers_architecture}), and the nine scalar predictions are concatenated to form the final coefficient vector. 

For SSL-Transformer, each M6 sample is treated as a compact coefficient snapshot rather than a spatial flow-field snapshot. 
The nine aerodynamic coefficients are arranged as nine tokens, each containing the normalized flow-condition information together with the normalized coefficient value. 
During training and evaluation, 20\% of the coefficient tokens are retained as observations and the model reconstructs the complete nine-coefficient vector. 
The encoder-decoder architecture is identical to that used in Experiment I; only the token structure differs, as summarised in Table~\ref{tab:burgers_architecture}.

Overall, the architecture comparison in Experiment II follows the same logic as Experiment I, but the task is now a sparse multi-output aerodynamic prediction problem. 
The GAN-based models learn the joint distribution of the input flow states and aerodynamic responses, RGFiL performs direct multi-target regression through independent gated experts, and SSL-Transformer performs token-based coefficient reconstruction from partial observations.

\subsubsection{Results and Analysis}

The prediction performance of the four compared models on the held-out M6 test set is summarized in Table~\ref{tab:m6_test_performance}. 
All metrics are calculated after inverse normalization, so the reported values correspond to the original aerodynamic-coefficient scale.

\begin{table}[!t]
  \caption{Test-set performance of the four compared models on the ONERA M6 dataset.}
  \label{tab:m6_test_performance}
  \centering
  \setlength{\tabcolsep}{4pt}
  \footnotesize
  \renewcommand{\arraystretch}{1.15}
  \begin{tabular*}{\textwidth}{@{\extracolsep{\fill}}cccccc@{}}
  \toprule
  Model & MAE & MSE & RMSE & $R^2$ & Corr. \\
  \midrule
  IG-GAN & $\mathbf{7.617\times10^{-4}}$ & $\mathbf{1.998\times10^{-6}}$ & $\mathbf{1.414\times10^{-3}}$ & $\mathbf{0.9998}$ & $\mathbf{0.9999}$ \\
  RBF-GAN & $1.402\times10^{-3}$ & $7.039\times10^{-6}$ & $2.653\times10^{-3}$ & $0.9994$ & $0.9997$ \\
  RGFiL & $1.206\times10^{-3}$ & $6.769\times10^{-6}$ & $2.602\times10^{-3}$ & $0.9994$ & $0.9997$ \\
  SSL-Transformer & $1.337\times10^{-3}$ & $1.172\times10^{-5}$ & $3.424\times10^{-3}$ & $0.9992$ & $0.9996$ \\
  \bottomrule
  \end{tabular*}
  \end{table}

IG-GAN achieves the best overall accuracy among the four models, with an MAE of $7.617\times10^{-4}$, an MSE of $1.998\times10^{-6}$, and an RMSE of $1.414\times10^{-3}$. 
Compared with RBF-GAN, IG-GAN reduces MAE, MSE, and RMSE by approximately $45.65\%$, $71.62\%$, and $46.70\%$, respectively. 
As in the Burgers' experiment, this gap is attributable to the generator alone, since both models use the identical RBF-D; the advantage is more pronounced here, which may simply reflect the smaller, sparser M6 training set rewarding a more constrained generator parameterization more heavily than a less-constrained one.
Compared with the RGFiL baseline, IG-GAN reduces MSE by approximately $70.48\%$, showing that the adversarial distribution-learning framework provides stronger accuracy than direct multi-output regression in this limited-data setting.

\begin{table*}[!t]
  \caption{Coefficient-wise test MSE of the four compared models on the ONERA M6 dataset.}
  \label{tab:m6_coefficient_mse}
  \centering
  \footnotesize
  \renewcommand{\arraystretch}{1.15}
  \begin{tabular*}{\textwidth}{@{\extracolsep{\fill}}ccccc@{}}
  \toprule
  Coefficient & IG-GAN & RBF-GAN & RGFiL & SSL-Transformer \\
  \midrule
  $C_L$ & $\mathbf{6.935\times10^{-6}}$ & $2.280\times10^{-5}$ & $1.692\times10^{-5}$ & $3.628\times10^{-5}$ \\
  $C_D$ & $\mathbf{4.627\times10^{-8}}$ & $1.324\times10^{-7}$ & $1.452\times10^{-7}$ & $1.862\times10^{-7}$ \\
  $C_{SF}$ & $\mathbf{2.785\times10^{-8}}$ & $4.688\times10^{-8}$ & $7.585\times10^{-8}$ & $4.364\times10^{-8}$ \\
  $C_{Mx}$ & $\mathbf{4.034\times10^{-6}}$ & $1.565\times10^{-5}$ & $1.770\times10^{-5}$ & $2.436\times10^{-5}$ \\
  $C_{My}$ & $\mathbf{1.016\times10^{-6}}$ & $2.249\times10^{-6}$ & $5.191\times10^{-6}$ & $6.156\times10^{-6}$ \\
  $C_{Mz}$ & $\mathbf{4.028\times10^{-8}}$ & $9.725\times10^{-8}$ & $9.572\times10^{-8}$ & $8.110\times10^{-8}$ \\
  $C_{Fx}$ & $\mathbf{4.041\times10^{-8}}$ & $4.245\times10^{-8}$ & $6.518\times10^{-8}$ & $5.411\times10^{-8}$ \\
  $C_{Fy}$ & $\mathbf{9.985\times10^{-9}}$ & $4.702\times10^{-8}$ & $5.276\times10^{-8}$ & $4.475\times10^{-8}$ \\
  $C_{Fz}$ & $\mathbf{5.836\times10^{-6}}$ & $2.228\times10^{-5}$ & $2.067\times10^{-5}$ & $3.828\times10^{-5}$ \\
  \bottomrule
  \end{tabular*}
  \end{table*}

The coefficient-wise errors in Table~\ref{tab:m6_coefficient_mse} further show that IG-GAN obtains the lowest MSE for all nine aerodynamic outputs. 
The improvement is particularly clear for $C_{Mx}$ and $C_{Fz}$, where the proposed model reduces the MSE by more than $70\%$ compared with RBF-GAN. 
This confirms that the overall advantage in Table~\ref{tab:m6_test_performance} is not caused by a single dominant coefficient, but is consistently observed across force, moment, and friction-related outputs.

\begin{figure}[!t]
  \centering
  \subfigure[$C_{Mx}$]{
      \includegraphics[width=0.96\textwidth]{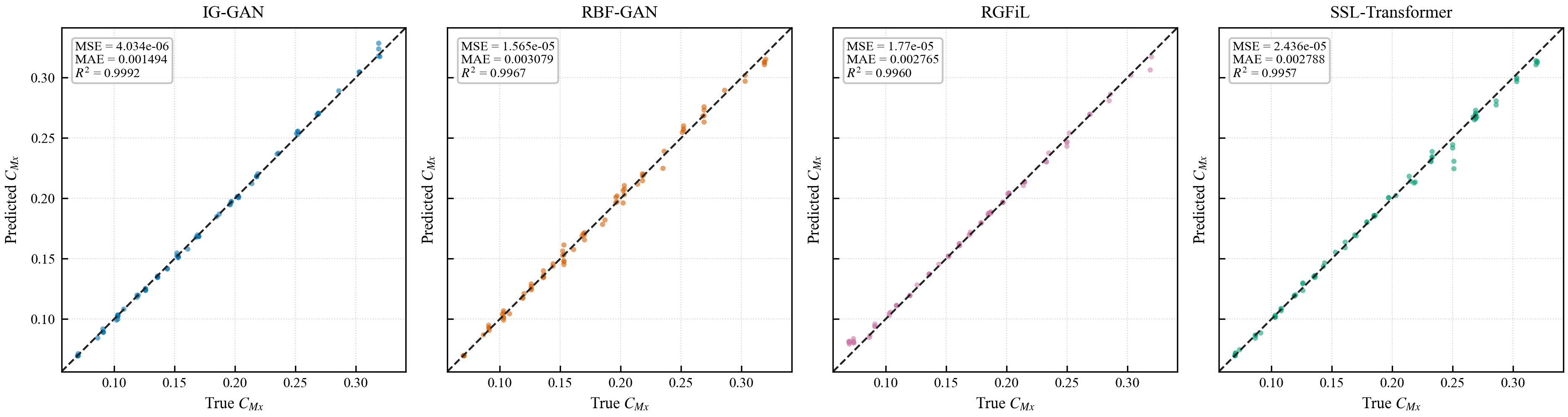}
  }
  \hfill
  \vspace{1mm}
  \subfigure[$C_{Fz}$]{
      \includegraphics[width=0.96\textwidth]{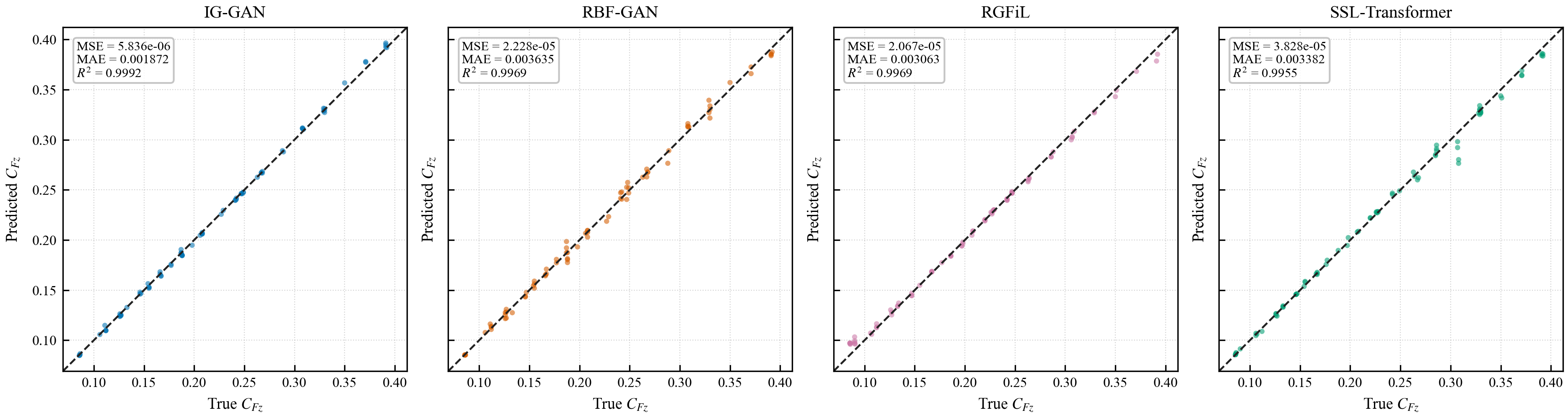}
  }
  \caption{Representative true-versus-predicted comparison scatter plots on the ONERA M6 test set. The diagonal dashed line denotes the ideal prediction.}
  \label{fig:m6_45degree_panel}
\end{figure}

The representative 45-degree scatter plots in Figure~\ref{fig:m6_45degree_panel} provide a direct comparison of the test-set prediction quality of the four models. 
Among the coefficient-wise comparison plots, $C_{Mx}$ and $C_{Fz}$ are selected because IG-GAN shows the clearest advantage on these two outputs, with the lowest error and the highest $R^2$ in both cases. 
The IG-GAN predictions are tightly distributed around the diagonal reference line, whereas the baseline models show slightly larger deviations, especially in the high-coefficient region. 
This indicates that the proposed model captures both the magnitude and the monotonic variation of the main aerodynamic responses.

\begin{figure}[!t]
  \centering
  \subfigure[$C_D$]{
      \includegraphics[width=0.4\textwidth]{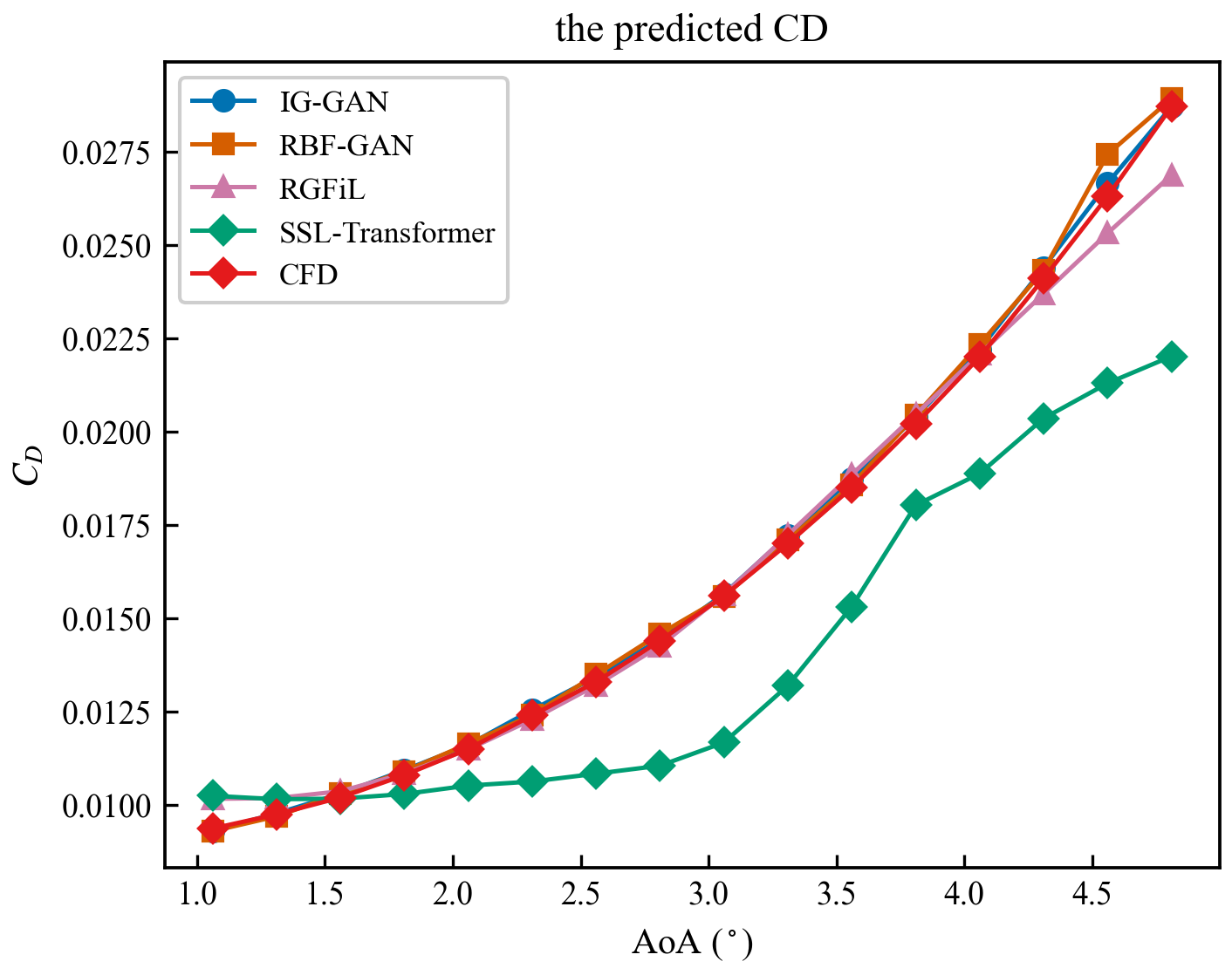}
  }
  \hfill
  \subfigure[$C_{Fy}$]{
      \includegraphics[width=0.4\textwidth]{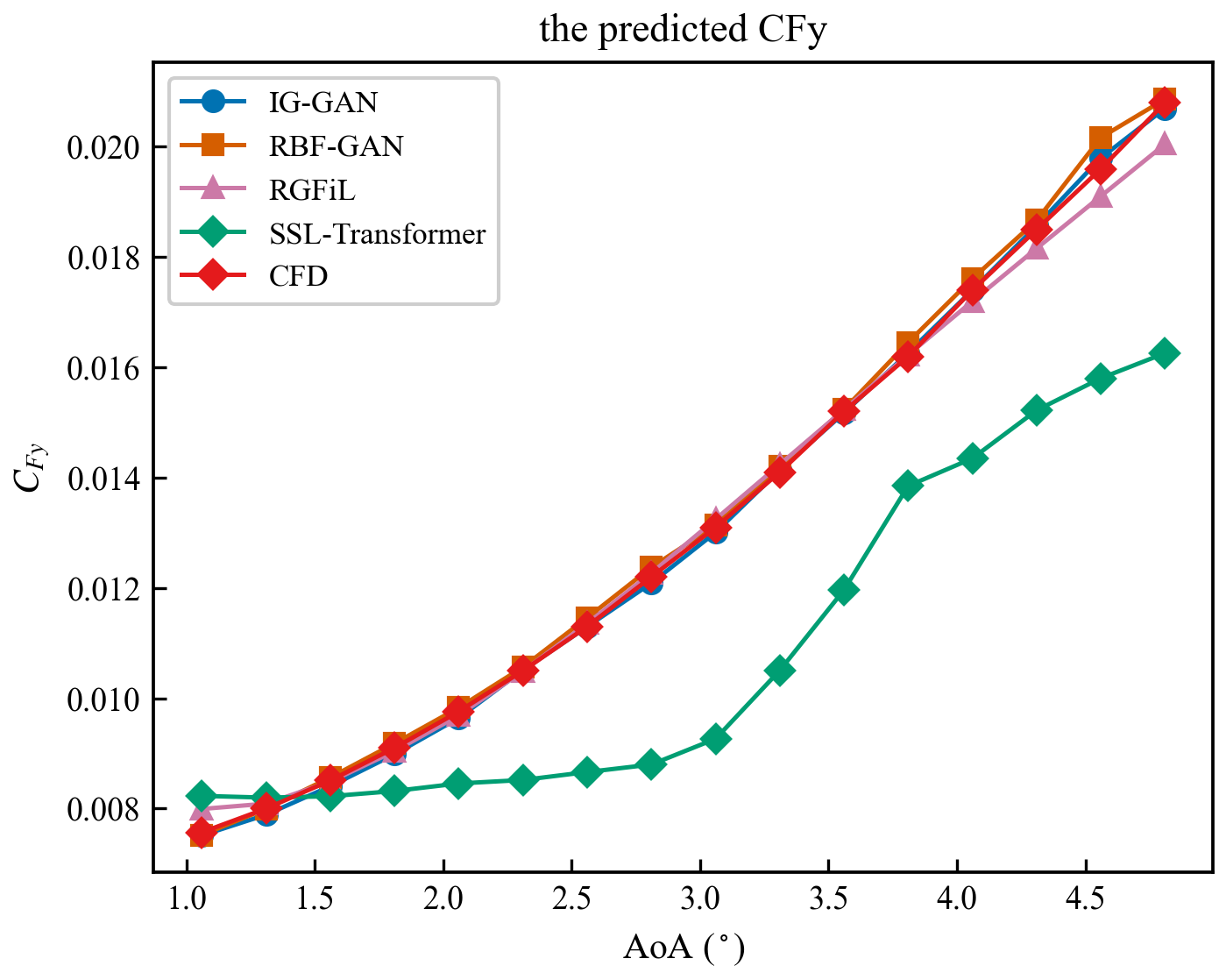}
  }
  \vspace{1mm}
  \subfigure[$C_{Mx}$]{
      \includegraphics[width=0.4\textwidth]{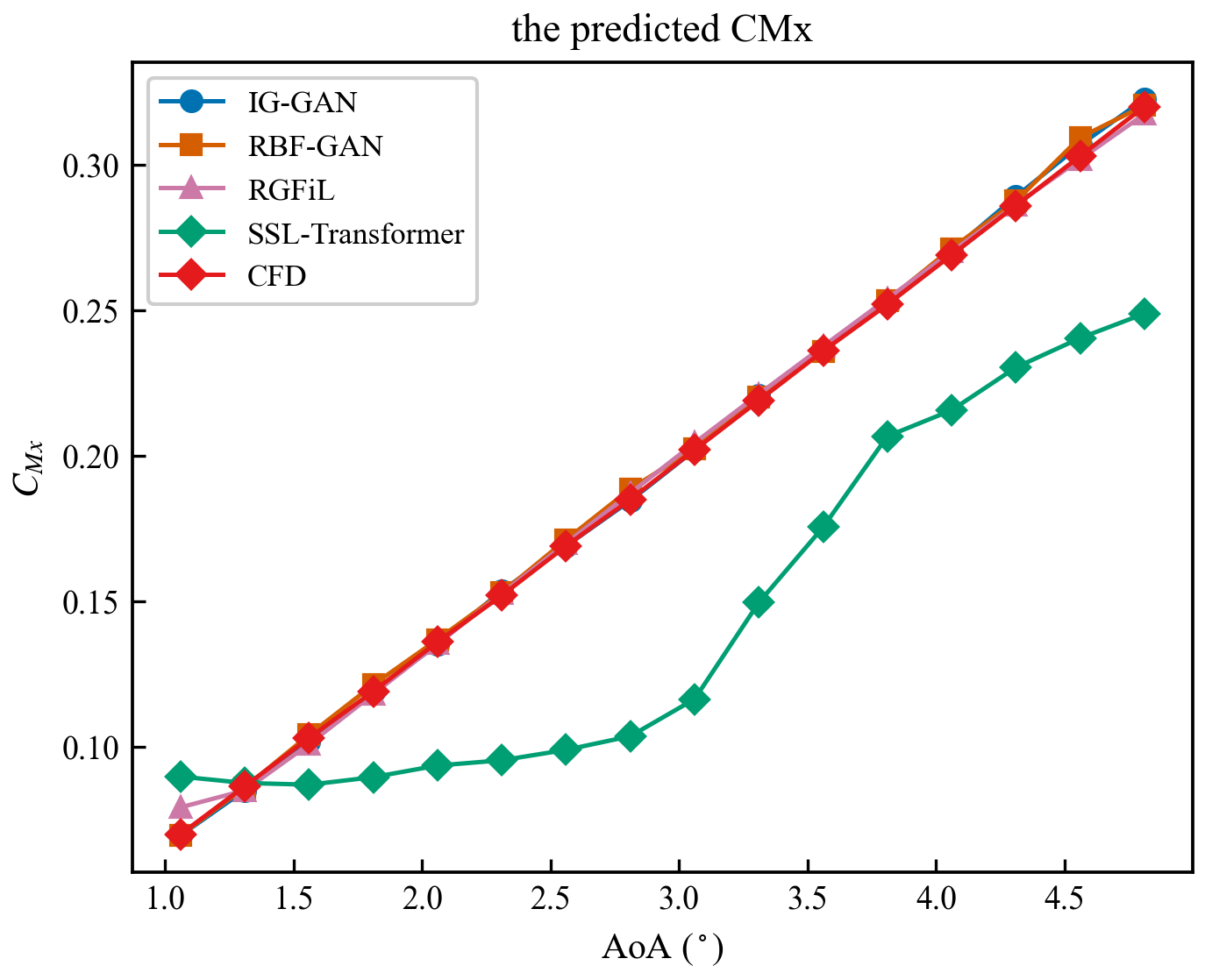}
  }
  \hfill
  \subfigure[$C_{SF}$]{
      \includegraphics[width=0.4\textwidth]{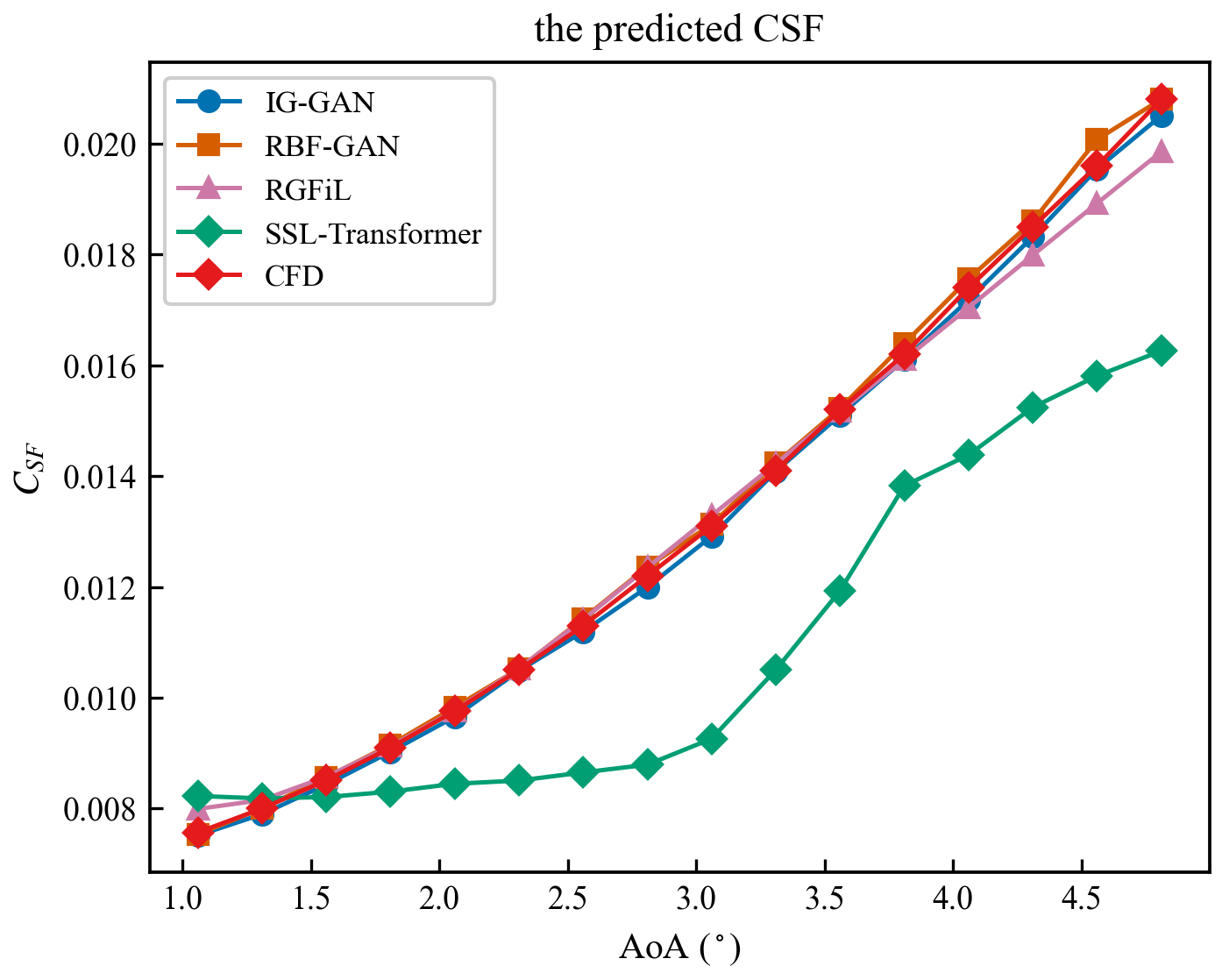}
  }
  \caption{Prediction curves with respect to angle of attack at $Ma=0.8045$ and $Re=1.2\times10^7$.}
  \label{fig:m6_aoa_curves}
\end{figure}

Figure~\ref{fig:m6_aoa_curves} presents the predicted aerodynamic coefficients with respect to the angle of attack at $Ma=0.8045$ and $Re=1.2\times10^7$. 
The CFD reference curves exhibit smooth monotonic variations over the investigated AoA range. 
Among all methods, IG-GAN achieves the closest agreement with the CFD results for all four coefficients. 
RBF-GAN yields comparable predictions with only minor discrepancies at higher angles of attack, whereas RGFiL slightly underestimates the coefficients in this region. 
By comparison, SSL-Transformer shows pronounced deviations in the medium- and high-AoA regimes, especially for $C_{Mx}$, indicating weaker extrapolation performance. 
Overall, the results confirm that IG-GAN provides the most accurate and robust predictions across different aerodynamic coefficients under this flow condition.

The M6 experiment confirms the advantage of IG-GAN in a sparse, nonlinear, multi-output aerodynamic prediction problem. 
The proposed model not only obtains the lowest global error, but also preserves physically meaningful coefficient trends with respect to angle of attack. 
This suggests that the intrinsic-geometry generator improves the representation of parameter-dependent aerodynamic responses, while the RBF discriminator helps constrain the generated samples in the joint input-output coefficient space.

\section{CONCLUSION}
\label{section_conclusion}

In this paper, we propose IG-GAN that incorporates a Bézier-surface-based intrinsic geometry generator and an RBF neural network discriminator to generate and predict nonlinear aerodynamic data. 
We demonstrate that aerodynamic data can be represented as a piecewise smooth manifold constructed from Bézier surfaces, and further demonstrate that the coefficient matrix of this manifold can be learned end-to-end by a neural network rather than solved in closed form. 
The experimental results demonstrate that IG-GAN achieves a $97.41\%$ reduction in test MSE for velocity reconstruction on the Burgers' dataset compared with SSL-Transformer, and an $82.95\%$ reduction in overall test MSE for aerodynamic coefficient prediction on the ONERA M6 dataset. 
Compared with RBF-GAN, which shares the same discriminator, IG-GAN reduces MSE by $51.19\%$ on the Burgers' dataset and $71.62\%$ on the ONERA M6 dataset, confirming that the Bézier-based generator is the source of this improvement. 
In practical applications, IG-GAN can be trained using historical flow-field or aerodynamic coefficient data, and the trained model can then be used to reconstruct flow fields or predict coefficients under new flight conditions, including conditions that are expensive to simulate directly.

In the future, we will further investigate the extrapolation ability of IG-GAN under unseen flow conditions and more complex aircraft geometries. 
In addition, extending the framework to higher-dimensional flow fields and integrating stronger physical constraints into the adversarial training process may further improve its robustness and applicability in practical aerodynamic design tasks.

\bibliographystyle{elsarticle-num}
\bibliography{main}

\end{document}